\documentclass[sigconf]{acmart}
\AtBeginDocument{%
  }

\setcopyright{CC}
\copyrightyear{2026}
\acmYear{2026}
\acmDOI{10.1145/3805622.3810805}
\acmConference[ICMR '26]{International Conference on Multimedia Retrieval}{June 16--19,
  2026}{Amsterdam, Netherlands}
\acmBooktitle{International Conference on Multimedia Retrieval (ICMR '26),
 June 16--19, 2026, Amsterdam, Netherlands}
\acmISBN{979-8-4007-2617-0/26/06}

\usepackage{graphicx}
\usepackage{subcaption}
\usepackage{multirow}

\begin{document}

\title{On the Effectiveness of Integration Methods for Multimodal Dialogue Response Retrieval}

\author{Seongbo Jang}
\orcid{0009-0008-3398-0144}
\affiliation{%
  \institution{Myongji University}
  \city{Seoul}
  \country{Republic of Korea}
}
\email{sjang@mju.ac.kr}

\author{Seonghyeon Lee}
\orcid{0009-0002-0039-8920}
\affiliation{%
  \institution{Kyungpook National University}
  \city{Daegu}
  \country{Republic of Korea}
}
\email{sh0416@knu.ac.kr}

\author{Dongha Lee}
\orcid{0000-0003-2173-3476}
\affiliation{%
  \institution{Yonsei University}
  \city{Seoul}
  \country{Republic of Korea}
}
\email{donalee@yonsei.ac.kr}

\author{Hwanjo Yu}
\authornote{Corresponding author.}
\orcid{0000-0002-7510-0255}
\affiliation{%
 \institution{Pohang University of Science and Technology}
 \city{Pohang}
 \country{Republic of Korea}
}
\email{hwanjoyupostech@gmail.com}


\begin{abstract}
  Multimodal chatbots have become one of the major topics for dialogue systems in both research community and industry.
Recently, researchers have shed light on the multimodality of responses as well as dialogue contexts.
This work explores how a dialogue system can output responses in various modalities such as text and image.
To this end, we first formulate a multimodal dialogue response retrieval task for retrieval-based systems as the combination of three subtasks.
We then propose three integration methods based on a two-step approach and an end-to-end approach, and compare the merits and demerits of each method.
Experimental results on two datasets demonstrate that the end-to-end approach achieves comparable performance without an intermediate step in the two-step approach.
In addition, a parameter sharing strategy not only reduces the number of parameters but also boosts performance by transferring knowledge across the subtasks and the modalities.
\end{abstract}




\maketitle

\section{Introduction}



As the demand for open-domain chatbots and technical development of dialogue systems rise steeply, researchers have brought attention to multimodal dialogue systems.
Among various modalities, image-grounded conversations have been actively researched along with the advent of several benchmarks~\citep{lin2014microsoft,Plummer_2015_ICCV,antol2015vqa,das2017visual,zellers2019recognition}.
Most of the benchmarks focus on the factual contents of images, usually given in the form of question-answer pairs.
In addition to factual information, recent studies started to consider emotional exchange and engagingness, which are humane aspects of open-domain dialogues~\citep{hu2014we}, by collecting more chit-chat-like datasets such as image-grounded conversation (IGC)~\citep{mostafazadeh-etal-2017-image} and ImageChat~\citep{shuster-etal-2020-image}.

While most work focused on understanding dialogue contexts of multiple modalities, little work tried to build an integrated system that outputs multimodal responses for retrieval-based chatbots.
\citet{zang-etal-2021-photochat} proposed photo-sharing intent prediction and image retrieval as individual tasks for multimodal response retrieval, yet the combination of an image retriever and a text retriever remains ambiguous.



To overcome this limitation, we first formulate multimodal response retrieval task which aims to choose the most appropriate text or image response for the next utterance, given a dialogue context composed of text utterances.
Then, we explore three unified methods that integrate subcomponents for the end task in different ways.
To be specific, dual retriever (DR) and shared dual retriever (SDR) are based on a two-step approach: 1) intent prediction determines the modality of the next utterance and 2) response retrieval finds out the most likely utterance from the predicted modality;
on the contrary, multimodal dual retriever (MDR) is an end-to-end approach that selects responses from a heterogeneous candidate pool of the both modalities without the explicit intent prediction step.
We emphasize that our contribution is orthogonal to cross-modal representation learning approaches~\citep{radford2021learning,akbari2021vatt}.
While these works share parameters to learn aligned representations from paired data under a single contrastive objective, we study how to \emph{integrate multiple retrieval subtasks of different modalities} into one dialogue system.



We evaluate the effectiveness of each method for multimodal response retrieval and investigate the effect of model size on two benchmark datasets.
The two-step approach performs better than the end-to-end approach in unimodal retrieval tasks, whereas the end-to-end approach achieves comparable performance compared to the two-step approach in multimodal retrieval, posing a question on the necessity of intent prediction.
In terms of model size, SDR and MDR reduce the number of parameters by sharing context encoders, which is impossible for DR that trains separate context encoders for each modality and intent prediction.

\section{Related Work}
  \begin{figure*}[ht]
    \centering
    \begin{subfigure}[b]{0.3\textwidth}
        \centering
        \includegraphics[width=\linewidth]{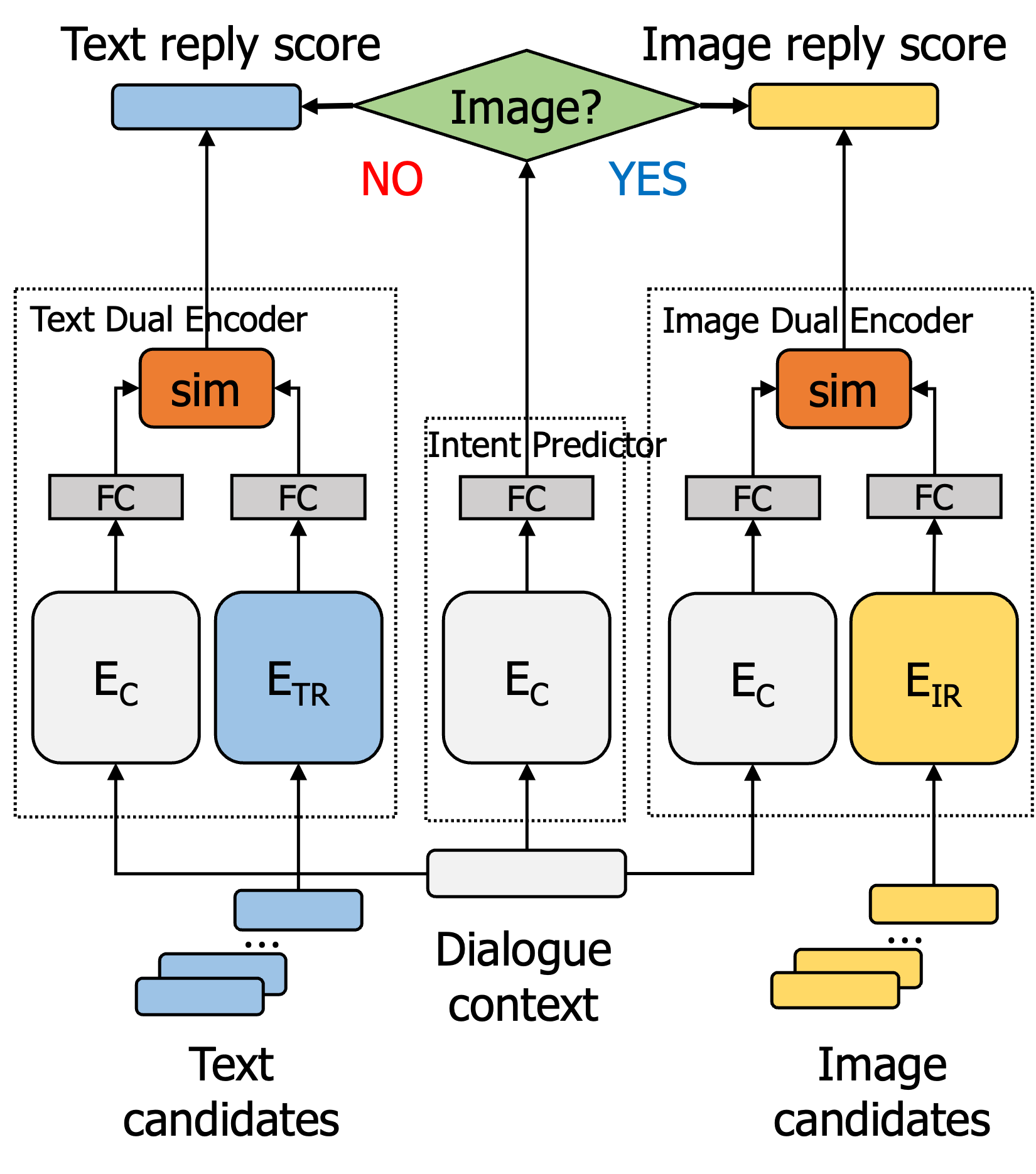}
        \caption{Dual Retriever}
        \Description{Dual Retriever}
    \end{subfigure}
    \hfill
    \begin{subfigure}[b]{0.3\textwidth}
        \centering
        \includegraphics[width=\linewidth]{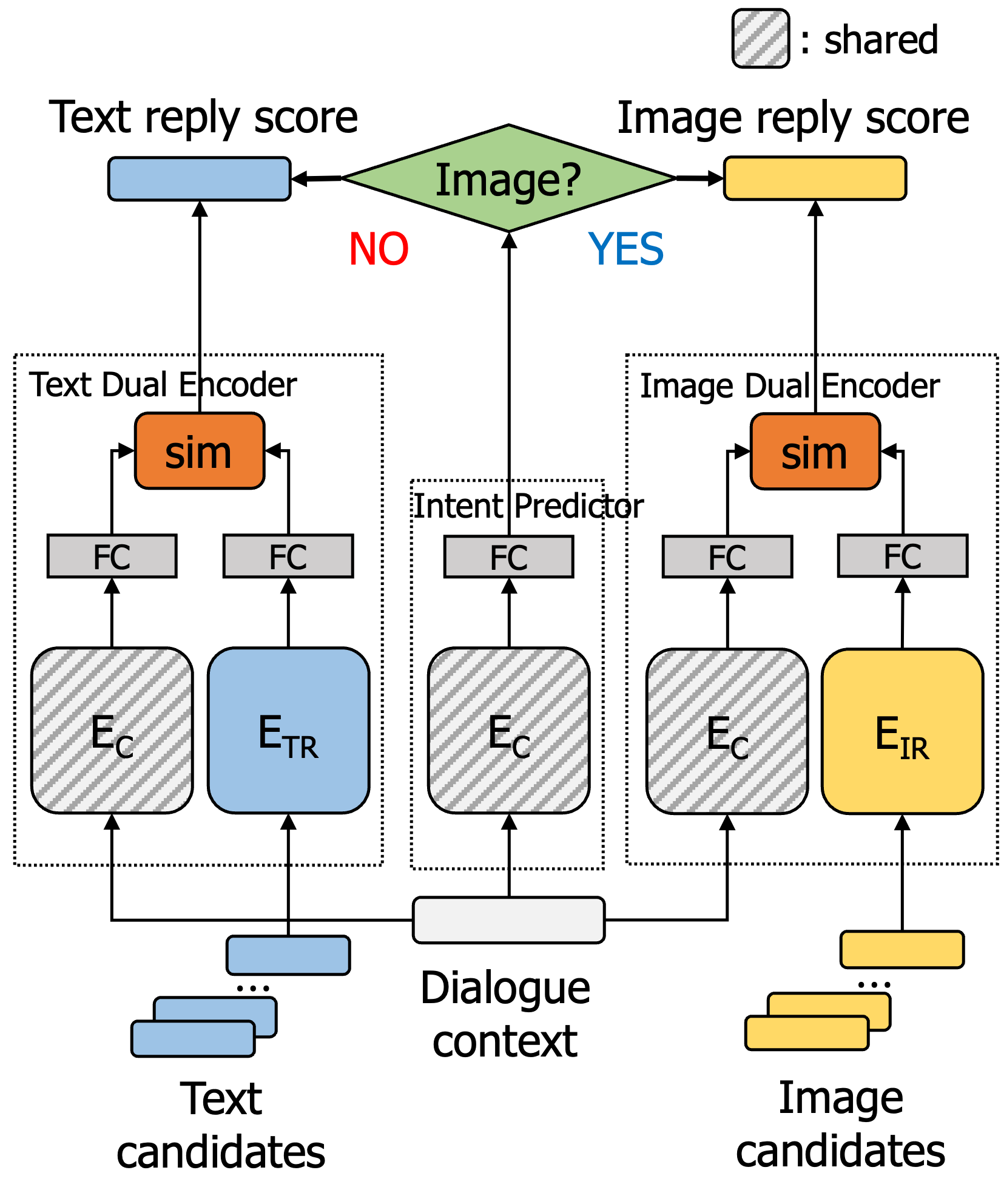}
        \caption{Shared Dual Retriever}
        \Description{Shared Dual Retriever}
    \end{subfigure}
    \hfill
    \begin{subfigure}[b]{0.3\textwidth}
        \centering
        \includegraphics[width=0.8\linewidth]{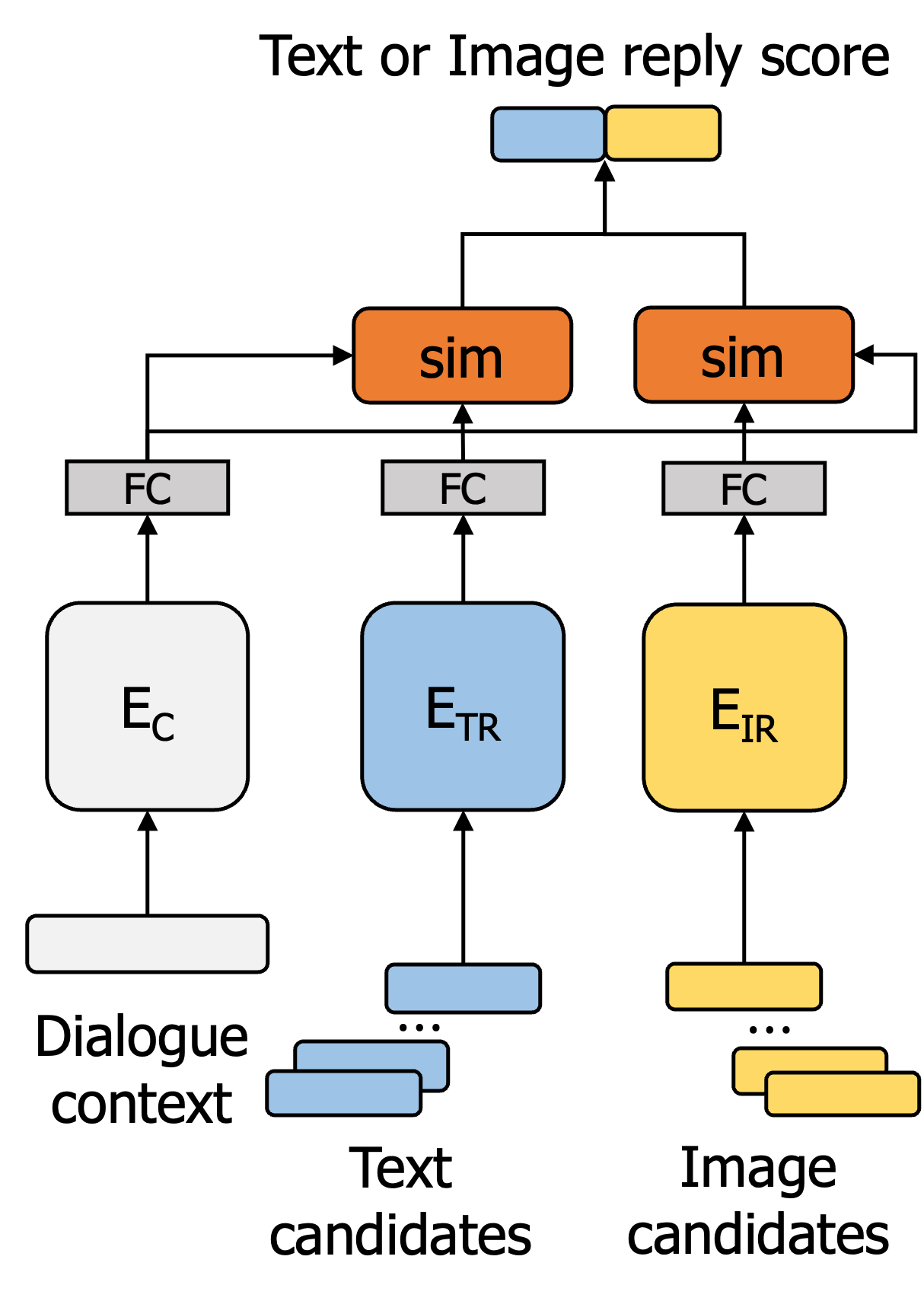}
        \caption{Multimodal Dual Retriever}
        \Description{Multimodal Dual Retriever}
    \end{subfigure}
    \caption{Three different architectures for the multimodal response retrieval task. $E_C$: context encoder, $E_{TR}$: text response encoder, $E_{IR}$: image response encoder. Unlike (a), the encoders in (b) share parameters across the three subtasks (denoted by the hatched pattern).}
    \Description{Overall Architecture}
    \label{fig:architectures}
\end{figure*}

\subsection{Dialogue Response Retrieval}




There are two representative approaches for dialogue response retrieval task: dual encoder and cross-encoder.
Dual encoder approaches~\citep{huang2013learning,henderson2017efficient,mazare-etal-2018-training,dinan2019wizard} encode dialogue contexts and responses separately with distinct encoders and compute matching scores to select the most appropriate response. 
This nature enables us to pre-compute representations to reduce computational burden at inference time, thus widely adopted in many studies~\citep{Humeau2020Poly-encoders:,lan2021exploring,chen2022contextual}.
In particular, researchers have focused on the selection of encoding functions such as CNNs~\citep{yan2018response,wu2018response}, RNNs~\citep{lowe-etal-2015-ubuntu}, combined architectures of CNNs and RNNs~\citep{yan2016learning,zhou-etal-2016-multi}, memory networks~\citep{zhang-etal-2018-personalizing}, transformers~\citep{dinan2019wizard,xu2021topic}, and BERT~\citep{henderson-etal-2019-training}.
Cross-encoder approaches~\citep{gu2020speaker,whang2020effective,wu-etal-2020-tod,Xu_Tao_Jiang_Zhao_Zhao_Yan_2021} take the concatenated inputs of contexts and responses to enrich interactions between the two.
They are also actively studied owing to the development of pre-trained language models (PLMs) such as BERT~\citep{devlin-etal-2019-bert}.
In this work, we focus on dual encoder architecture since cross-encoders are less preferred to dual encoders in practical situations.

\subsection{Multimodal Open-domain Dialogue}





There have been several studies to build multimodal open-domain dialogue systems which output text responses related to a given image followed by a few turns of dialogue contexts.
\citet{shuster-etal-2020-image} constructed ImageChat dataset which involves images and dialogue contexts along with personality traits allocated to speakers, and proposed a unified architecture using Transformer~\citep{vaswani2017attention} and ResNet~\citep{7780459}.
\citet{shuster-etal-2020-dialogue} built a multi-task dialogue agent using 12 open-domain dialogue datasets including ImageChat and IGC.
On the success of Blender~\citep{roller-etal-2021-recipes}, \citet{shuster-etal-2021-multi} incorporated an image encoder~\citep{xie2017aggregated} to enable image-grounded conversation.
\citet{sun2022multimodal} focused on generation models which generate either text or image responses conditioned on the preceding textual dialogue contexts.
In contrast, we study the retrieval setting and propose, to the best of our knowledge, propose the first integrated system for multimodal dialogue response retrieval.

\section{Method}

\subsection{Task: Multimodal Response Retrieval}
Multimodal response retrieval aims to select the most appropriate response among the text candidates and image candidates for a given dialogue context.
The text is represented as a sequence of tokens $r^\mathbf{t} = (t_1, ..., t_l)$ where each token is included in a pre-defined vocabulary.
The image is represented by 3-dimensional tensor $r^\mathbf{i} \in \mathbb{R}^{H \times W \times C}$.
To solve this task, the model predicts the score $s(c, r)$ that indicates how much each response $r$ in the candidate pool is appropriate for a dialogue context $c = (c_1, ..., c_l)$ where $c_i$ is previous text utterance.
Using this score, the model selects the response with the highest score for the given context.
Therefore, the goal of the model is to accurately select the ground truth response while harmonizing the multimodal response candidates,
\begin{equation}
    r^* = \underset{{r \in \{r^\mathbf{t}_1,...,r^\mathbf{t}_i\} \cup \{r^\mathbf{i}_1,...,r^\mathbf{i}_j\}}}{\mathrm{argmax}} s(c, r). \nonumber
\end{equation}

\paragraph{Subtask: Intent Prediction} aims to determine the modality of the next utterance given a context.
In the case of two modalities (text and image), an intent prediction model $f_i$ takes a context $c$ as input and produces a binary logit.
Given the pair of context and either image response or text response, the binary cross entropy loss is used for training as follows:
\begin{equation*}
\resizebox{0.8\linewidth}{!}{%
$
\mathcal{L}_{\text{intent}} = \sum_{(c, r^i)} \mathcal{L}_\mathrm{BCE} ( f_i(c), 1 ) + \sum_{(c, r^t)} \mathcal{L}_\mathrm{BCE} ( f_i(c), 0).$%
}
\end{equation*}

\paragraph{Subtask: Text Response Retrieval} is to select the most appropriate text response given the current context.
We adopt dual encoder architecture widely used in response selection~\citep{yang-etal-2018-learning}.
To be specific, a context encoder $f_c^\mathbf{t}$ and a response encoder $f_r^\mathbf{t}$ compute the representations of a context $c$ and a response $r$ respectively.
The cosine similarity between two representations is regarded as the score, and the encoders are optimized to accurately predict the score based on the cross entropy loss.
The loss computed from $i$-th pair of context $c_i^\mathbf{t}$ and text response $r^\mathbf{t}_i$ is as follows:
\begin{align*}
    \mathcal{L}_\mathrm{text} =& - \log \frac{\exp(s_{\mathbf{t},\mathbf{t}}(c_i^\mathbf{t},r^\mathbf{t}_i ))}{\sum_{(\cdot, r^\mathbf{t}_j) \in B} \exp(s_{\mathbf{t},\mathbf{t}}(c_i^\mathbf{t}, r^\mathbf{t}_j))} \\
    & - \log \frac{\exp(s_{\mathbf{t},\mathbf{t}}(c_i^\mathbf{t},r^\mathbf{t}_i))}{\sum_{(c^\mathbf{t}_j, \cdot) \in B} \exp(s_{\mathbf{t},\mathbf{t}}(c_j^\mathbf{t}, r^\mathbf{t}_i))},
\end{align*}
where $s_{\cdot,\cdot}(c,r)=\cos ( f_c^\cdot (c), f_r^\cdot (r))$ for further notational simplicity.
The dot ($\cdot$) in the summation indices denotes a placeholder that is marginalized over all candidates in the batch $B$.

\paragraph{Subtask: Image Response Retrieval} handles image responses during conversation.
This task is the same with text response retrieval except that the modality of response is image.
Therefore, the response encoder is built on the pretrained image encoder \citep{7780459}.
The loss for the $i$-th pair of context $c_i^\mathbf{t}$ and image response $r^\mathbf{i}_i$ is described as follows:
\begin{align*}
    \mathcal{L}_\mathrm{image} =& - \log \frac{\exp(s_{\mathbf{t},\mathbf{i}}(c_i^\mathbf{t}, r^\mathbf{i}_i))}{\sum_{(\cdot, r^\mathbf{i}_j) \in B} \exp(s_{\mathbf{t},\mathbf{i}}(c_i^\mathbf{t}, r^\mathbf{i}_j))} \\
    & - \log \frac{\exp(s_{\mathbf{t},\mathbf{i}}(c_i^\mathbf{t},r^\mathbf{i}_i))}{\sum_{(c^\mathbf{t}_j, \cdot) \in B} \exp(s_{\mathbf{t},\mathbf{i}}(c_j^\mathbf{t}, r^\mathbf{i}_i))}.
\end{align*}

While the models for each subtask are trained separately, these models do not capture useful supervision across the subtasks and cannot effectively solve the multimodal response retrieval task.
Considering the knowledge derived across the subtasks, we introduce three approaches (Fig.~\ref{fig:architectures}) to integrate these modules in the following subsections.

\subsection{Dual Retriever (DR)}
One simple integration is to weave the separately trained models.
Each model is trained by three different optimization problems:
\begin{gather*}
\resizebox{0.8\linewidth}{!}{%
$
    \underset{\theta_i}{\mathrm{minimize}}\;\mathcal{L}_\mathrm{intent}\text{, }
    \underset{\theta_c^\mathbf{t}, \theta_r^\mathbf{t}}{\mathrm{minimize}}\;\mathcal{L}_\mathrm{text}\text{, }
    \underset{\theta_c^\mathbf{t}, \theta_r^\mathbf{i}}{\mathrm{minimize}}\;\mathcal{L}_\mathrm{image}\text{,}$%
}
\end{gather*}
where $\theta$ represents the parameters of each model.
Using these trained models, DR combines the produced outputs to obtain final outputs during inference.
To be specific, the intent predictor predicts the modality of the response for an input context.
Depending on the prediction, we select the corresponding retrieval model and then find out the most appropriate response.
\begin{equation*}
\resizebox{0.7\linewidth}{!}{%
$
    f(c^\mathbf{t}) = 
    \begin{cases}
        \underset{{r^\mathbf{i}_{i}}}{\mathrm{argmax}}\quad s_{\mathbf{t},\mathbf{i}}(c^\mathbf{t}, r^\mathbf{i}_i) & \text{if } f_i(c) > 0.5,\\
        \underset{{r^\mathbf{t}_{i}}}{\mathrm{argmax}}\quad s_{\mathbf{t},\mathbf{t}}(c^\mathbf{t}, r^\mathbf{t}_i) & \text{otherwise.}
    \end{cases}$%
}
\end{equation*}
However, since the supervision from each modality is not transferred across different retrievers, the derived knowledge is not fully reflected to both retrieval models.

\subsection{Shared Dual Retriever (SDR)}
We propose a simple but effective scheme that shares the retriever to encourage the models to communicate with each other across the subtasks.
Although the architecture of response encoder is different due to the different modality, the architecture of context encoder is the same.
Thus, we can share the parameters of the context encoder between the two subtasks.
With the help of parameter sharing, we integrate the optimization problems for two response retrieval tasks:
\begin{gather*}
    \underset{\theta_i}{\mathrm{minimize}}\;\mathcal{L}_\mathrm{intent}\text{, }
    \underset{\theta_{c}^\mathbf{t}, \theta_r^\mathbf{t}, \theta_r^\mathbf{i}}{\mathrm{minimize}}\;\mathcal{L}_\mathrm{text}+\mathcal{L}_\mathrm{image}.
\end{gather*}

Furthermore, we make the context encoder inside the intent predictor ($\theta_i = \theta_c$) share its parameters with those of the response retrieval models.
By doing so, the separate optimization problems are merged into a unified optimization problem as follows:
\begin{equation*}
    \underset{\theta_{c}^\mathbf{t}, \theta_r^\mathbf{t}, \theta_r^\mathbf{i}}{\mathrm{minimize}}\;\mathcal{L}_\mathrm{intent} + \mathcal{L}_\mathrm{image} + \mathcal{L}_\mathrm{text}.
\end{equation*}

We further hypothesize the ineffectiveness inside the inference process due to the intent predictor.
One reason is the cascaded error coming from the intent predictor.
The intent predictor acts as a branch for choosing the modality of the most appropriate response.
However, it means that the final prediction result is wrong if the intent predictor predicts wrong modality.
In addition, from the recent success in modeling cross-modal representation space \citep{radford2021learning,akbari2021vatt}, we hypothesize that the response representation space from different modalities can become naturally aligned.

\subsection{Multimodal Dual Retriever (MDR)}
From the above hypothesis, we propose our final integration approach by removing the intent predictor and modeling multimodal response representation space.
We remove the intent predictor and directly compare the cosine similarities across different modality.
Then an integrated response encoder is defined as $f_r^\mathbf{m} = f_r^\mathbf{i}$ if $r=r^\mathbf{i}$, or $f_r^\mathbf{m} = f_r^\mathbf{t}$ if $r=r^\mathbf{t}$.
The loss for the $i$-th pair of the two different modalities is defined as follows:
\begin{align*}
    \mathcal{L}_\mathrm{joint} =& - \log \frac{\exp(s_{\mathbf{t},\mathbf{m}}(c^\mathbf{t}_i, r_i))}{\sum_{(\cdot ,r_j) \in B} \exp(s_{\mathbf{t},\mathbf{m}}(c^\mathbf{t}_i, r_j))} \\
    & - \log \frac{\exp(s_{\mathbf{t},\mathbf{m}}(c^\mathbf{t}_i, r_i))}{\sum_{(c_j, \cdot) \in B} \exp(s_{\mathbf{t},\mathbf{m}}(c^\mathbf{t}_j, r_i))},
\end{align*}
where a batch $B$ consists of context-response pairs whose response is either image or text.
All the parameters can be effectively optimized to minimize the joint loss in an end-to-end manner,
\begin{equation*}
    \underset{\theta_{c}, \theta_{r^\mathbf{t}}, \theta_{r^\mathbf{i}}}{\mathrm{minimize}}\;\mathcal{L}_{\mathrm{joint}}.
\end{equation*}
Note that the absence of the intent predictor further simplifies the inference process by taking the response that has the most highest score among all the multimodal candidates.


\section{Experiments}
  \begin{table*}[ht]
\centering
\caption{Results on the test set of PhotoChat.
Text (image) retrieval: retrieval among 50 text (image) response candidates for the examples whose ground truths are text (image) responses, assuming the intent predictor is an oracle for DR and SDR. MDR does not get advantage in this setting since there is no explicit intent prediction step.
Small: $\text{BERT}_{\textsc{Mini}}$ and $\text{ResNet}_{50}$, Large: $\text{BERT}_{\textsc{Base}}$ and $\text{ResNet}_{152}$~\citep{turc2019well}.}
\begin{tabular}{@{}clccccccccc@{}}
\toprule
\multicolumn{1}{l}{}   &                            & \multicolumn{3}{c}{Text Retrieval}                                                    & \multicolumn{3}{c}{Image Retrieval}                                                   & \multicolumn{3}{c}{Multimodal Retrieval}                                              \\ \midrule
Model Size             & \multicolumn{1}{c}{Method} & \multicolumn{1}{c}{R@1} & \multicolumn{1}{c}{R@5} & \multicolumn{1}{c}{R@10} & \multicolumn{1}{c}{R@1} & \multicolumn{1}{c}{R@5} & \multicolumn{1}{c}{R@10} & \multicolumn{1}{c}{R@1} & \multicolumn{1}{c}{R@5} & \multicolumn{1}{c}{R@10} \\ \midrule
\multirow{3}{*}{Small} & DR                         & \textbf{0.4325}            & 0.7371                     & 0.8701                      & \textbf{0.4519}            & \textbf{0.7994}            & \textbf{0.9007}             & 0.2475                     & 0.4234                     & 0.4949                      \\
                      & SDR                        & 0.4274                     & \textbf{0.7553}            & \textbf{0.8721}             & 0.4396                     & 0.7787                     & 0.8873                      & \textbf{0.4000}            & \textbf{0.7046}            & \textbf{0.8086}             \\
                      & MDR                        & 0.3990                     & 0.7076                     & 0.8518                      & 0.4122                     & 0.7269                     & 0.8599                      & 0.3766                     & 0.6594                     & 0.7721                      \\ \midrule
\multirow{3}{*}{Large}  & DR                         & 0.4305                     & 0.7482                     & 0.8751                      & \textbf{0.4620}            & \textbf{0.8247}            & \textbf{0.9058}             & 0.2079                     & 0.3560                     & 0.4143                      \\
                      & SDR                        & \textbf{0.4650}            & \textbf{0.7990}            & \textbf{0.9066}             & 0.4315                     & 0.8061                     & 0.9046                      & \textbf{0.4152}            & \textbf{0.7239}            & 0.8157                      \\
                      & MDR                        & 0.4315                     & 0.7299                     & 0.8812                      & 0.4406                     & 0.7797                     & 0.8964                      & 0.4020                     & 0.6949                     & \textbf{0.8193}             \\ \bottomrule
\end{tabular}%
\label{tab:photochat}
\end{table*}

\begin{table*}[htbp]
\centering
\caption{Results on the test set of MMDial. Defined terms are the same as in Table~\ref{tab:photochat}.}
\begin{tabular}{@{}clccccccccc@{}}
\toprule
\multicolumn{1}{l}{}   &                            & \multicolumn{3}{c}{Text Retrieval}                  & \multicolumn{3}{c}{Image Retrieval}                 & \multicolumn{3}{c}{Multimodal Retrieval}            \\ \midrule
Model Size             & \multicolumn{1}{c}{Method} & R@1             & R@5             & R@10            & R@1             & R@5             & R@10            & R@1             & R@5             & R@10            \\ \midrule
\multirow{3}{*}{Small} & DR                         & 0.6118          & 0.8039          & \textbf{0.8946} & 0.0430          & 0.2148          & 0.3687          & \textbf{0.2874} & 0.3749          & 0.4161          \\
                       & SDR                        & \textbf{0.6138} & \textbf{0.8115} & 0.8870          & \textbf{0.1356} & \textbf{0.3958} & \textbf{0.5430} & 0.2619          & 0.3934          & 0.4527          \\
                       & MDR                        & 0.5386          & 0.7558          & 0.8512          & 0.1281          & 0.3846          & 0.5334          & 0.2731          & \textbf{0.4517} & \textbf{0.5378} \\ \midrule
\multirow{3}{*}{Large} & DR                         & \textbf{0.6333} & \textbf{0.8453} & \textbf{0.9141} & 0.0605          & 0.2204          & 0.3544          & 0.2900          & 0.3835          & 0.4125          \\
                       & SDR                        & 0.5986          & 0.8313          & 0.8986          & \textbf{0.1484} & \textbf{0.4300} & \textbf{0.6002} & 0.2516          & 0.4113          & 0.4735          \\
                       & MDR                        & 0.5967          & 0.8202          & 0.8938          & 0.1281          & 0.4097          & 0.5728          & \textbf{0.3079} & \textbf{0.4883} & \textbf{0.5895} \\ \bottomrule
\end{tabular}%
\label{tab:mmdial}
\end{table*}

We elucidate the datasets, evaluation metric used in our experiments, and implementation details in Appendix~\ref{appendix}.

\subsection{Effects of Integration Methods}

In contrast to relatively small gaps of R@$k$ among the three methods in unimodal retrieval, the end task performance is largely affected depending on how we integrate the subcomponents.
On PhotoChat (Table~\ref{tab:photochat}), SDR achieves the highest R@$k$ for multimodal retrieval, except that MDR reaches slightly higher R@10 when the model size is large.
Meanwhile on MMDial (Table~\ref{tab:mmdial}), MDR achieves the highest R@$k$ for multimodal retrieval, except that DR scores the highest R@1 for the small model.
We note that although DR outperforms MDR in text retrieval and image retrieval, MDR outperforms DR on the contrary due to the cascaded error from the intent prediction step.
These results highlight the significance of choosing an appropriate integration method of submodules.

\subsection{Effects of Model Size}

Overall, all the methods in large models tend to achieve higher R@$k$ than their counterparts in small models on the both datasets.
On PhotoChat, MDR shows comparable performance for multimodal retrieval to SDR when the model size grows large.
Similarly on MMDial, MDR effectively increases R@$k$ for multimodal retrieval compared to the two-step approach when the large model is used.
These results suggest that large models are more effective at aligning multimodal response representations, which benefits end-to-end approaches such as MDR.

\subsection{Effects of Parameter Sharing}

On both datasets, the performance of DR for multimodal retrieval lags behind those of SDR and MDR which share the parameters of context encoders.
DR trains each submodule separately on the three individual subtasks so none of the subcomponents can get the knowledge from the other subtasks and modalities positively transferred, resulting in disharmony for accomplishing the ultimate goal~\citep{wu2021fixes}.

In addition, weight sharing decreases the number of total parameters from 72M to 49M in small models and from 501M to 281M in large models, which become around 1.5x and 1.8x smaller respectively.

\section{Conclusion}


We propose an integrated task to build a multimodal dialogue system that outputs both text and image responses, and present three architectures named DR, SDR, and MDR.
We then empirically analyze the effectiveness of intent prediction which was introduced in the previous work.

The experimental results on two datasets demonstrate that the end-to-end approach without an intent predictor is competitive with the two-step approach for multimodal retrieval.
In addition, SDR and MDR successfully reduce the number of model parameters without compromising the end task performance.
\bibliographystyle{ACM-Reference-Format}
\bibliography{anthology,custom}

\appendix

\section{Experimental Setup}
  \label{appendix}
  \subsection{Datasets}



\paragraph{PhotoChat~\citep{zang-etal-2021-photochat}} is a multimodal dialogue dataset which includes dyadic dialogues covering various topics in daily lives.
It consists of 10,286/1,000/1,000 dialogue contexts with a single image attached to each context in the train/dev/test set.
There exist 8,889/1,000/1,000 unique images along with object labels in the train/dev/test set.

\paragraph{Multi-modal Dialogue (MMDial)~\citep{lee-etal-2021-constructing}} was constructed by substituting text utterances in existing text-only dialogue datasets with relevant images from large-scale image datasets using a state-of-the-art image-text matching model~\citep{Li_2019_ICCV}.
It consists of 39,956/2,401/2,673 dialogue contexts with a single image attached to each context in the train/dev/test set.
There exist 12,272/334/682 unique images in the train/dev/test set.

\paragraph{Preprocessing.}

For each dialogue context, we extract the first $n \in \{1, 2, \ldots, l\}$ utterances as an example to augment contexts, where $l$ is the length of a context before sharing an image.
Since object labels are not explicitly attached in MMDial, we instead use the original image captions from the source datasets (MS-COCO and Flickr30k) as object labels for each image.
While such caption-based matching could introduce noise, the captions in MS-COCO and Flickr30k are human-annotated, so the resulting matches are reliable in practice.

\subsection{Evaluation Metric}




We measure recall at $k$ (R@$k$) to evaluate multimodal response retrieval. It calculates the occurrence when the gold response is retrieved in top-$k$ ($k=1,5,10$) candidates.
For dev and test, each query uses a fixed pool of 50 candidates per modality, randomly sampled from all images and text responses in the dev and test sets, respectively, and reused across all models for fair comparison.
We use the checkpoint that achieves the best dev R@$k$ and report the test R@$k$ evaluated by the checkpoint.

\subsection{Implementation Details}

\paragraph{Architectural details of encoders}

The context encoder ($E_C$) and the text response encoder ($E_{TR}$) consist of a single BERT encoder followed by a projection layer.
In all three architectures, the context encoder ($E_C$) and the text response encoder ($E_{TR}$) share parameters, since we observed this consistently yielded better performance than keeping them separate.
Beyond this sharing, SDR additionally shares the context encoder across the intent and image-retrieval subtasks.
The image response encoder ($E_{IR}$) consists of an image encoder which extracts visual features with ResNet and a object label encoder which extracts object label features with BERT.
Note that both BERT and ResNet are used to encode image features since an object label is attached to every image.
The two representations are then concatenated and projected to the joint embedding space.
We use $\text{BERT}_{\textsc{Mini}}$ and $\text{ResNet}_{50}$ for small models and $\text{BERT}_{\textsc{Base}}$ and $\text{ResNet}_{152}$ for large models as specified by~\citet{turc2019well}.

\paragraph{Hyperparameters}

For BERT, the dropout rate is 0.2 and the maximum sequence length is set to 128.
We use cosine similarity with the temperature $\tau = 0.01$ as the similarity measure.
We train the model for 10 epochs for intent prediction and text dialogue retrieval, and 20 epochs for image dialogue retrieval and multimodal dialogue retrieval with the batch size of 64 in one V100 GPU.
We apply Adam optimizer~\citep{kingma2015adam} with the learning rate of 5e-5 and linear decay of 0.1\% per 1,000 steps.

\end{document}